\documentclass[10pt,letterpaper]{article}
\usepackage[top=0.85in,left=2.75in,footskip=0.75in]{geometry}
\usepackage{amsmath,amssymb}
\usepackage{changepage}
\usepackage{textcomp,marvosym}
\usepackage{cite}
\usepackage{nameref,hyperref}
\usepackage[right]{lineno}
\usepackage[nopatch=eqnum]{microtype}
\DisableLigatures[f]{encoding = *, family = * }
\usepackage[table]{xcolor}
\usepackage{array}
\usepackage{multirow}
\usepackage{adjustbox}
\usepackage{array}
\usepackage{tabularx}
\usepackage{subcaption}

\newcolumntype{+}{!{\vrule width 2pt}}

\newlength\savedwidth

\usepackage{setspace} 
\raggedright
\usepackage[aboveskip=1pt,labelfont=bf,labelsep=period,justification=raggedright,singlelinecheck=off]{caption}
\renewcommand{\figurename}{Fig}
\makeatletter
\renewcommand{\@biblabel}[1]{\quad#1.}
\makeatother

\usepackage{lastpage,fancyhdr,graphicx}
\usepackage{epstopdf}
\fancyheadoffset[L]{2.25in}
\fancyfootoffset[L]{2.25in}
\begin{document}
\vspace*{0.2in}

\begin{flushleft}
{\Large
\textbf\newline{Beyond accuracy: quantifying the reliability of Multiple Instance Learning for Whole Slide Image classification} 
}
\newline

Hassan Keshvarikhojasteh\textsuperscript{1*},
Marc Aubreville\textsuperscript{2},
Christof A. Bertram\textsuperscript{3},
Josien P.W. Pluim\textsuperscript{1},
Mitko Veta\textsuperscript{1}
\\
\bigskip
\textbf{1} Department of Biomedical Engineering, Eindhoven University of Technology, Eindhoven, The Netherlands
\\
\textbf{2} Flensburg Artificial Intelligence Research Group (FLAIR), Flensburg University of Applied Sciences, Flensburg, Germany
\\
\textbf{3} University of Veterinary Medicine Vienna, Vienna, Austria
\\
\bigskip

* h.keshvarikhojasteh@tue.nl

\end{flushleft}

\section*{Abstract}
Machine learning models have become integral to many fields, but their reliability, defined as producing dependable, trustworthy, and domain-consistent predictions, remains a critical concern. Multiple Instance Learning (MIL) models designed for Whole Slide Image (WSI) classification in computational pathology are rarely evaluated in terms of reliability, leaving a key gap in understanding their suitability for high-stakes applications like clinical decision-making. In this paper, we address this gap by introducing three quantitative metrics for reliability assessment and applying them to several widely used MIL architectures across three region-wise annotated pathology datasets. Our findings indicate that the mean pooling instance (MEAN-POOL-INS) model demonstrates superior reliability compared to other networks, despite its simple architectural design and computational efficiency. These findings underscore the need of reliability evaluation alongside predictive performance in MIL models and establish MEAN-POOL-INS as a strong, trustworthy baseline for future research.

\section*{Introduction}
Machine learning (ML) has become a cornerstone of modern computational pathology by enabling automated analysis of large and complex datasets generated in clinical practice \cite{esteva2017dermatologist, litjens2017survey}. Deep learning, in particular, has shown remarkable potential for tasks such as tumor detection, tissue subtyping, and prognosis prediction \cite{coudray2018classification, campanella2019clinical}. 

Multiple Instance Learning (MIL) is a widely used method for weakly-supervised classification tasks in computational pathology, primarily due to the difficulty and labor-intensive nature of obtaining pixel-wise annotations from histopathology data \cite{dietterich1997solving, carboneau2018milSurvey}. Compared to simpler approaches such as patch-level classifiers with propagated slide labels or heuristic aggregation strategies, MIL offers a principled means of reducing label noise and focusing on diagnostically relevant regions. In the MIL framework, groups of instances, or "bags," are created, with the goal of predicting the label for each bag. The key assumption in MIL is that positive bags contain at least one key instance, while negative bags do not. Recent advancements in deep learning have spurred the development of novel MIL models tailored to computational pathology,  enhancing performance through key instance detection \cite{ilse2018attention}, feature space refinement \cite{lu2021data}, and overfitting mitigation \cite{zhang2022dtfd, zh2023acmil}.

Reliability refers to the ability of a system or model to perform as expected under standard and well-defined conditions. The reliability of MIL models is essential for gaining the trust of physicians, ensuring successful deployment in clinical settings, and adhering to established healthcare standards. Reliable models focus on biologically pertinent features, grounded in established scientific knowledge, when making predictions. In the context of computational pathology, an unreliable MIL model might erroneously focus on artifacts or irrelevant features, leading to false positives or negatives in critical tasks such as tumor detection or subtyping.  For example, a model could achieve high slide-level accuracy by correctly predicting tumor presence in most cases, yet consistently rely on irrelevant regions such as staining or scanner-induced artifacts, making it accurate but not reliable. Conversely, a model could consistently focus on biologically meaningful regions (reliable) but exhibit slightly lower overall accuracy due to ambiguous or noisy labels. Such errors could result in misdiagnoses, delays in treatment, or unnecessary interventions, underscoring the need for robust reliability evaluations. Achieving reliability in MIL models for computational patholopgy is particularly challenging due to the inherent variability in histopathology data, including differences in staining, scanning resolutions, and patient-specific biological heterogeneity \cite{komura2018machine}. Furthermore, the weakly-supervised nature of MIL exacerbates these challenges by relying on coarse slide-level labels without precise spatial annotations. Addressing these limitations requires not only model improvements but also careful evaluation frameworks that ensure predictions align with domain knowledge and clinical relevance. By fostering reliable MIL models, computational pathology can advance toward broader acceptance and integration into clinical workflows.

However, machine learning models are rarely evaluated in terms of reliability; instead, performance metrics such as accuracy and F1 score are typically reported \cite{kelly2019key}. Yet, reliable models are essential for real-world deployment and tend to exhibit better generalization. Therefore, gaining insights into models' reliability is critical. For instance, Pendyala et al. \cite{pendyala2024assessing} assessed the reliability of various models using a mental health dataset and discovered that these models often focused on less relevant features, despite achieving high performance, thereby highlighting the limitations of traditional metrics. Furthermore, Araujo et al. \cite{araujo2023quest} propose an alternative validation method for models and support the assertion that performance metrics are not indicative of the reliability of machine learning networks. In the domain of medical imaging, the trustworthiness of saliency maps has been quantitatively assessed on two public radiology datasets using the area under the precision-recall curve (AUPRC) and the structural similarity index measure (SSIM) as evaluation metrics \cite{arun2021assessing}. The study revealed that none of the saliency maps satisfied all the tested criteria, highlighting limitations in their reliability. 

Interpretability and reliability in the context of machine learning (ML) are related but distinct concepts that address different aspects of model performance and trustworthiness. Interpretability means how easily a person can understand why a model made a certain decision or how it comes to its results \cite{biran2017explanation}. We aim to understand and explain the way models work in interpretability, providing transparency. To address questions concerning the decision-making processes and the underlying mechanics, numerous interpretability methods have been developed \cite{salahuddin2022transparency}.  Interpretability and explainability methods have proven useful in identifying both data and model deficiencies \cite{javed2022additive}. For MIL models, attention scores assigned to each patch are frequently used to generate heatmaps, which are key tools for assessing interpretability. However, interpretability in MIL models has traditionally been evaluated qualitatively by presenting specific slides and their corresponding heatmaps. The Camelyon16 dataset \cite{bejnordi2017diagnostic}, which includes pixel-wise ground truth annotations, is the most common resource for such evaluations. While qualitative assessments provide valuable insights, they are not feasible across large test sets due to computational and time constraints. Moreover, analyzing heatmaps requires specialized histopathology knowledge, which may not be readily available to machine learning researchers. Finally, a more robust and comprehensive assessment  should involve evaluation across multiple datasets. 

In contrast to interpretability, reliability is concerned with ensuring the model consistently produces correct and trustworthy predictions in a wide range of circumstances, independent of transparency of underlying process by which predictions are made. Models focusing on related regions of interest (ROIs) are more reliable, as they process the most relevant parts of the data. However, more interpretable models are not necessarily more reliable. For instance, a model might offer more interpretability through class-specific information or excitatory-inhibitory heatmaps \cite{javed2022additive}, yet fail to concentrate on relevant ROIs. Additionally, while interpretable models may provide explanations for predictions across entire classes, this does not necessarily align with reliability. For example, in binary tumor versus normal classification, since normal samples lack clear ROIs, predictions for these cases are not considered in reliability assessments.

In this paper, we address the above-mentioned challenges by quantitatively evaluating the reliability of MIL models for both binary tumor versus normal classification and multi-class tissue subtyping problems using three metrics. Additionally, we incorporate a recently proposed additive interpretability method \cite{javed2022additive} to investigate whether improved interpretability translates into increased reliability, providing new insights into the complex relationship between interpretability and reliability. Unlike previous studies that rely mainly on single-dataset evaluations, we extend our experiments beyond the Camelyon16 dataset by including two additional datasets: CATCH \cite{wilm2022pan} and TCGA BRCA. By introducing quantitative reliability metrics, benchmarking multiple MIL architectures across three datasets, and examining the impact of additive interpretability, this study provides a comprehensive, multi-dataset evaluation of MIL model reliability in computational pathology. Our study aims to advance the development of more reliable  MIL models for computational pathology.

\bigskip

\noindent The main contributions of this paper are as follows:
\begin{itemize}
	\item Introduction of three quantitative metrics to evaluate the reliability of MIL models in computational pathology.
	
	\item Systematic benchmarking of several widely-used MIL architectures across three region-wise annotated pathology datasets: Camelyon16, CATCH, and TCGA BRCA.
	
	\item Investigation of the impact of a recently proposed additive interpretability method on MIL model reliability, providing insights into the relationship between interpretability and reliability.
	
	\item Demonstration that comprehensive reliability evaluation complements traditional accuracy metrics and is essential for trustworthy deployment in high-stakes clinical settings.
\end{itemize}

\section*{Materials and methods}
In this section, we describe the problem formulation, the evaluation metrics for predictive performance and reliability, the datasets used in our experiments, and the details of our implementation.

\subsection*{Problem formulation}
\begin{figure}[!ht]
	\caption{ \textbf{The overall framework for evaluating the reliability of MIL models follows a three-step process.} First, a MIL model is trained on a weakly-supervised task for predicting slide-level labels. Next, the trained model is applied to predict scores for individual image patches. Finally, the reliability value is computed based on the predicted patch scores and their corresponding annotations, where tumor patches are highlighted in green and normal patches in orange in the annotation visualization.}
	\label{fig:anno}
	\includegraphics[width=\textwidth]{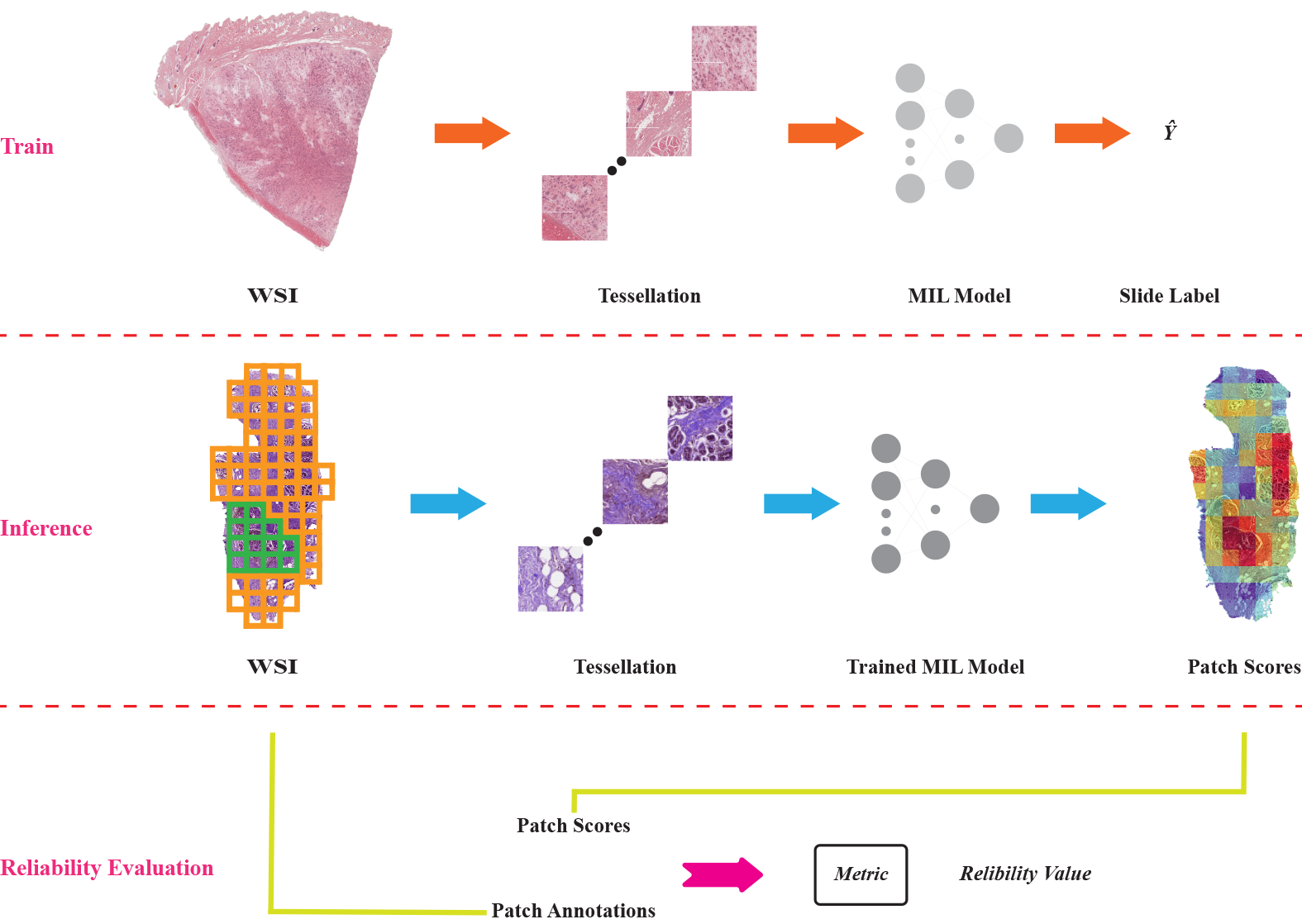}
\end{figure}

For our analysis, we refer to the input Whole Slide Image (WSI) as $I$ and predict its slide-level label $\hat{Y}$.  Following established methodologies, we first tessellate the slide into $N$ small patches and then extract features from these patches using a pre-trained model. In the final step, the slide label is predicted using an aggregation module, which is typically attention-based, providing attention scores for each patch. For clarification, the proposed framework is illustrated in \figurename~\ref{fig:anno}. 

\subsection*{Metrics}

Reliability in this study is defined as the consistent focus of MIL models on diagnostically relevant ROIs within WSIs, a prerequisite for trustworthy and clinically useful predictions. To quantify this alignment, we selected three complementary metrics that together capture different aspects of spatial concordance between predicted patch scores and ground truth annotations:

\begin{itemize}
	\item \textbf{Mutual Information $(MI)$:} Measures the statistical dependence between predicted scores and ROIs, quantifying how much information the model’s output shares with the true ROI distribution \cite{ross2014mutual}.

	\item \textbf{Spearman’s Correlation $(Spearman\text{’}s)$:} Captures the monotonic relationship between predicted patch importance and true ROI presence, allowing for non-linear associations and reflecting relative ranking fidelity \cite{spearman1904proof}.
	
	\item \textbf{Area Under the Precision-Recall Curve $(AUPRC)$:} Evaluates the model’s ability to localize ROIs under conditions of class imbalance by measuring precision and recall trade-offs \cite{saito2015precision}.

\end{itemize} 

\subsection*{Dataset}
Three public WSI datasets were used in the experiments. The Breast Invasive Carcinoma dataset (\textbf{TCGA BRCA}) from The Cancer Genome Atlas (TCGA) project includes 1,038 H\&E-stained slides representing two types of breast cancer: Invasive Ductal Carcinoma (IDC) and Invasive Lobular Carcinoma (ILC), with slide-level labels. The \textbf{CAMELYON16} dataset contains 399 WSIs with breast cancer lymph node metastasis annotations, divided into 270 slides for training and validation, and 129 for testing. Finally, the \textbf{CATCH} dataset provides 350 meticulously annotated slides from seven different canine cutaneous tumors.

Unlike CAMELYON16 and CATCH, the TCGA BRCA dataset lacks ground truth region-level annotations. To address this, tumor regions in the test slides were manually annotated with the assistance of Dr. Christof A. Bertram, a pathologist \cite{banerjee2025comprehensive}.

\subsection*{Implementation details}
Throughout the experiments, we employ the Adam optimizer with a learning rate of \(1 \times 10^{-4}\) and set the maximum number of epochs to 50. The weight decay is tuned based on the validation loss. All experiments are repeated with five different random seeds, and we report the mean \textit{AUC} and \textit{F1} scores as classification metrics, along with the mean \textit{MI}, \textit{$r_s$}, and \textit{AUPRC} values as reliability metrics. The \textit{AUC} measures the model’s overall discriminative ability in classifying positive and negative cases across all thresholds, while the \textit{AUPRC} emphasizes the model’s reliability in identifying true positive regions under conditions of class imbalance. Using \textit{AUPRC} as a reliability metric thus complements the classification-based \textit{AUC}, providing a more comprehensive evaluation of MIL model performance. Additionally, we evaluate \textit{model size} and \textit{Floating-Point Operations Per Second (FLOPS)} as computational metrics. For dataset preprocessing, we follow the procedures outlined in \cite{lu2021data}, which involve extracting features from non-overlapping \(256 \times 256\) patches at \(20\times\) magnification using a ResNet50 pretrained on ImageNet. These extracted features are fed into the model with an input dimensionality of $input\_dim = 1024$, which is subsequently reduced to $D = 512$.

We evaluate several state-of-the-art models, including ABMIL \cite{ilse2018attention}, CLAM \cite{lu2021data}, DTFD \cite{zhang2022dtfd}, ACMIL \cite{zh2023acmil}, MADMIL \cite{keshvarikhojasteh2024multi}, mean pooling (MEAN-POOL), max pooling (MAX-POOL), max pooling instance (MAX-POOL-INS), and mean pooling instance (MEAN-POOL-INS). For the MEAN-POOL and MAX-POOL models, we calculate the slide representation after feature compression using either the mean or max operation, and then predict the slide label. It should be noted that reliability metrics are not reported for the MEAN-POOL model, as its mean aggregation does not yield patch-level scores or attention weights. In contrast, for the MAX-POOL-INS and MEAN-POOL-INS models, we first predict patch probabilities, which are then aggregated to generate slide-level probabilities. Patch probabilities are also used to compute reliability metrics for MEAN-POOL-INS. In the case of MAX-POOL-INS, we use both the selected patch index, similar to MAX-POOL, and the patch probabilities.

For multi-head models such as MADMIL and ACMIL, average attention scores or patch probabilities are employed to calculate reliability metrics. We use a shorthand notation to indicate the number of heads, e.g., MADMIL/4 refers to the MADMIL model with four heads. To ensure a thorough analysis, we also investigate additive versions of these models, in accordance with the approach proposed by \cite{javed2022additive} , which reformulates the final predictor as an additive function of individual instances. For additive models, either attention scores or patch probabilities are used to assess reliability.

To compute the reliability metrics, it is essential to obtain both patch labels and the corresponding attention scores or probabilities. Following tessellation, patch labels are assigned based on the region-wise ground truth annotations. Specifically, for each patch, if it falls within a annotated region , it is assigned a label of 1, indicating the presence of the target feature. Conversely, if the patch does not intersect with the annotated region, it is assigned a label of 0, denoting the absence of the target feature. 

\section*{Results}

We present the predictive performance and reliability results for CAMELYON16, CATCH, and TCGA BRCA datasets. This section highlights how well each model identifies regions of interest and compares their overall effectiveness.  

\subsubsection*{CAMELYON16}
\begin{figure}[!ht]
	\caption{\textbf{(I) The test-30 slide with ground truth annotations (green) overlaid on the tissue section. (II) Corresponding heatmap generated by MAX-POOL, showing predicted patch scores distribution from low (blue) to high (red).} The annotation and heatmap are spatially aligned for comparison.}
	\label{fig:cam_max}
	\includegraphics[width=\textwidth]{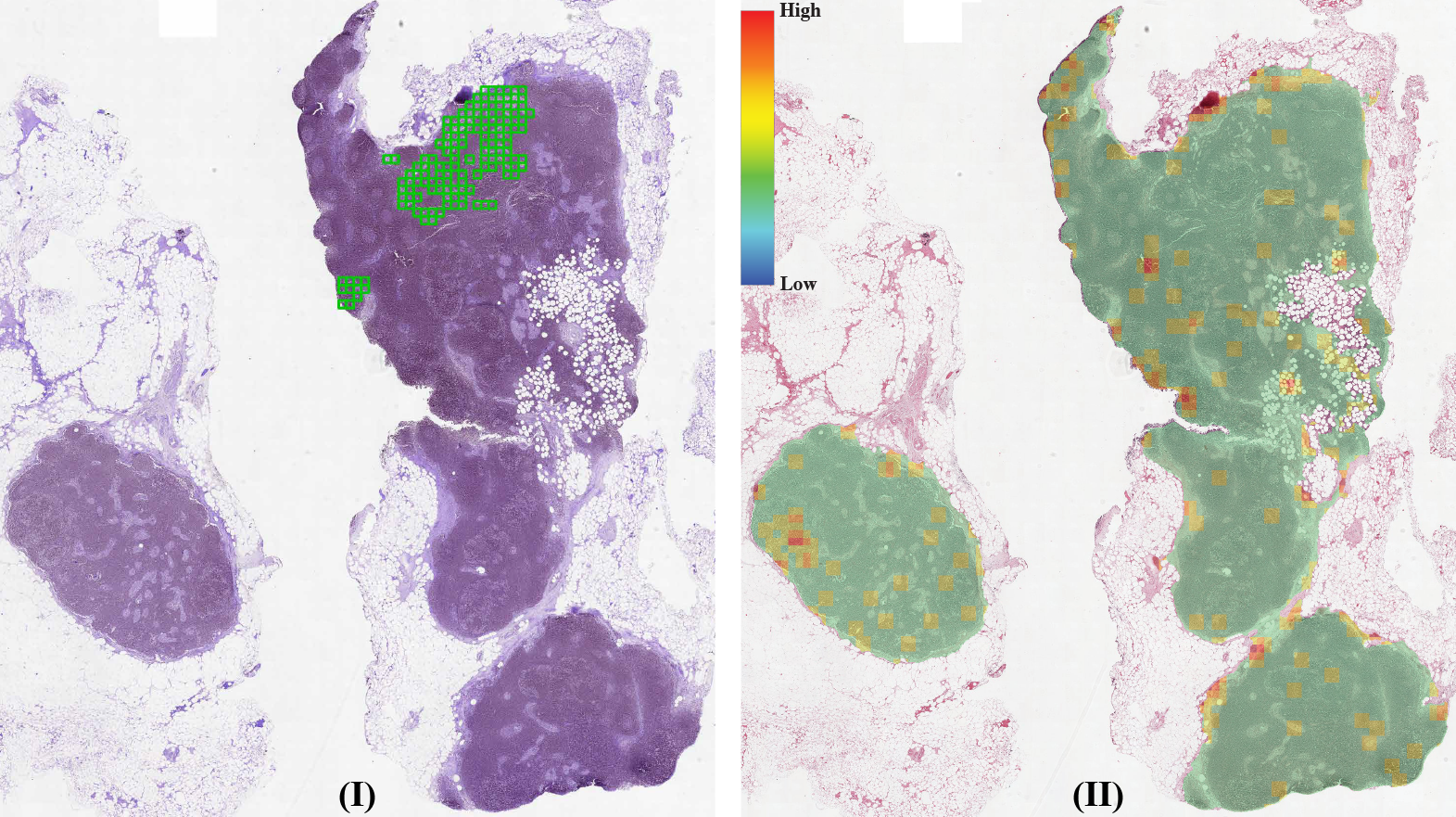}
\end{figure}

The results for this dataset are presented in Tables ~\ref{table1} and~\ref{table2}. In Table ~\ref{table1}, the ACMIL/2 model demonstrates moderate  classification performance but high reliability. Conversely, DTFD achieves strong classification performance with lower reliability. Interestingly, the MAX-POOL model, while computationally efficient and yielding reasonable classification results, exhibits the lowest reliability. This finding reinforces our argument that model selection should not be based solely on classification metrics. To illustrate this point, we present a tumor slide that was correctly classified by MAX-POOL, along with its heatmap in \figurename~\ref{fig:cam_max}. The heatmap reveals that the model concentrates on normal patches  of lymph node tissue rather than metastatic tumor patches, indicating the unreliability of its predictions.

\begin{table}[!ht]
	\begin{adjustwidth}{-0.74in}{0in} 
		\centering
		\renewcommand{\arraystretch}{2} 
		
		\scriptsize
		\caption{\bf Average reliability, classification, and computational metrics (± standard deviation) over five repetitions for the CAMELYON16 dataset.}
		\begin{tabular}{|l|c|c|c|c|c|c|c|}
			\hline
			\multirow{2}{*}{\bf Model} & \multicolumn{3}{c|}{\bf Reliability} & \multicolumn{2}{c|}{\bf Classification} & \multicolumn{2}{c|}{\bf Computation} \\
			\cline{2-8}
			& {\bf MI} & {\bf Spearman's} & {\bf AUPRC} & {\bf AUC} & {\bf F1} & {\bf FLOPs} & {\bf Size} \\
			\hline
			ABMIL & 0.12 ± 0.00 & 0.29 ± 0.00 & 0.59 ± 0.01 & 0.85 ± 0.01 & 0.84 ± 0.01 & 94 M & 789 K \\
			\hline
			\bf CLAM & 0.12 ± 0.01 & 0.29 ± 0.02 & 0.59 ± 0.02 & 0.86 ± 0.01 & \bf 0.85 ± 0.01 & 94 M & 791 K \\
			\hline
			\bf DTFD & 0.13 ± 0.01 & 0.29 ± 0.01 & 0.61 ± 0.01 & \bf 0.89 ± 0.01 & 0.85 ± 0.02 & 126 M & 1053 K \\
			\hline
			\bf ACMIL/2 & \bf 0.16 ± 0.02 & \bf 0.34 ± 0.02 & 0.59 ± 0.05 & 0.82 ± 0.03 & 0.81 ± 0.01 & 94 M & 791 K \\
			\hline
			\bf MADMIL/2 & 0.13 ± 0.00 & 0.31 ± 0.00 & \bf 0.61 ± 0.00 & 0.86 ± 0.01 & 0.85 ± 0.01 & 79 M & 658 K \\
			\hline
			\bf MAX-POOL & 0.00 ± 0.00 & -0.00 ± 0.00 & 0.23 ± 0.01 & 0.80 ± 0.01 & 0.79 ± 0.01 & \bf 63 M & \bf 526 K \\
			\hline
			MEAN-POOL\textsuperscript{a} & – & – & – & 0.58 ± 0.02 & 0.59 ± 0.02 & 63 M & 526 K \\
			\hline
		\end{tabular}
		
		\begin{flushleft}
			FLOPs are computed with 120 instances per bag. The highest value in each column is bolded using the exact value, while the tables report numbers rounded to two decimals for readability.\\
			\textbf{a} Reliability metrics are not computed for this model, as it does not provide patch scores.
		\end{flushleft}
		\label{table1}
	\end{adjustwidth}
\end{table}

\begin{table}[!ht]
	\begin{adjustwidth}{-1.27in}{0in} 
		\centering
		\renewcommand{\arraystretch}{2} 
		
		\scriptsize
		\caption{
			{\bf Average reliability, classification, and computational metrics (± standard deviation) over five repetitions for the CAMELYON16 dataset using additive models.}}
		\begin{tabular}{|l|c|c|c|c|c|c|c|}
			\hline
			\multirow{2}{*}{\bf Model} & \multicolumn{3}{c|}{\bf Reliability} & \multicolumn{2}{c|}{\bf Classification} & \multicolumn{2}{c|}{\bf Computation} \\
			\cline{2-8}
			& {\bf MI} & {\bf Spearman's} & {\bf AUPRC} & {\bf AUC} & {\bf F1} & {\bf FLOPs} & {\bf Size} \\
			\hline
			ABMIL-ADD-ATT & 0.13 ± 0.00 & 0.32 ± 0.01 & 0.62 ± 0.01 & \multirow{2}{*}{0.86 ± 0.01} & \multirow{2}{*}{0.83 ± 0.02} & \multirow{2}{*}{95 M} & \multirow{2}{*}{789 K} \\
			\cline{1-4}
			ABMIL-ADD-PATCH & 0.04 ± 0.00 & 0.24 ± 0.01 & 0.48 ± 0.01 &&&& \\
			\hline
			CLAM-ADD-ATT & 0.12 ± 0.01 & 0.29 ± 0.01 & 0.60 ± 0.02 & \multirow{2}{*}{0.87 ± 0.01} & \multirow{2}{*}{0.83 ± 0.02} & \multirow{2}{*}{95 M} & \multirow{2}{*}{791 K} \\
			\cline{1-4}
			CLAM-ADD-PATCH & 0.03 ± 0.00 & 0.24 ± 0.05 & 0.48 ± 0.02 &&&& \\
			\hline
			\bf DTFD-PATCH\textsuperscript{a} & 0.04 ± 0.00 & 0.29 ± 0.02 & 0.53 ± 0.01 & \bf 0.89 ± 0.01 & \bf 0.85 ± 0.02 & 126 M & 1053 K \\
			\hline
			ACMIL/2-ADD-ATT & 0.17 ± 0.01 & 0.35 ± 0.01 & 0.62 ± 0.02 &  \multirow{2}{*}{0.79 ± 0.02} &  \multirow{2}{*}{0.81 ± 0.01} &  \multirow{2}{*}{95 M} &  \multirow{2}{*}{791 K} \\
			\cline{1-4}
			ACMIL/2-ADD-PATCH & 0.03 ± 0.00 & -0.03 ± 0.01 & 0.29 ± 0.01 &&&& \\
			\hline
			MAX-POOL-INS & 0.00 ± 0.00 & 0.05 ± 0.03 & 0.20 ± 0.02 & \multirow{2}{*}{0.77 ± 0.14} & \multirow{2}{*}{0.76 ± 0.13} & \multirow{2}{*}{63 M} & \multirow{2}{*}{526 K} \\
			\cline{1-4}
			MAX-POOL-INS-PATCH & 0.15 ± 0.06 & 0.29 ± 0.15 & 0.55 ± 0.20 &&&& \\
			\hline
			\bf MEAN-POOL-INS & \bf 0.31 ± 0.04 & \bf 0.53 ± 0.07 & \bf 0.78 ± 0.08 & 0.58 ± 0.02 & 0.59 ± 0.02 & \bf 63 M & \bf 526 K \\
			\hline
		\end{tabular}
		
		\begin{flushleft}
			\textbf{a} We use the provided patch probabilities based on GRAD-CAM.
		\end{flushleft}
		
		\label{table2}
	\end{adjustwidth}
\end{table}

\begin{figure}[!ht]
	\caption{\textbf{(I) The test-40 slide with ground truth annotations (green) overlaid on the tissue section. (II) Corresponding heatmap generated by MEAN-POOL-INS, showing predicted patch scores distribution from low (blue) to high (red).} The annotation and heatmap are spatially aligned for comparison.}
	\label{fig:cam_mean}
	\includegraphics[width=\textwidth]{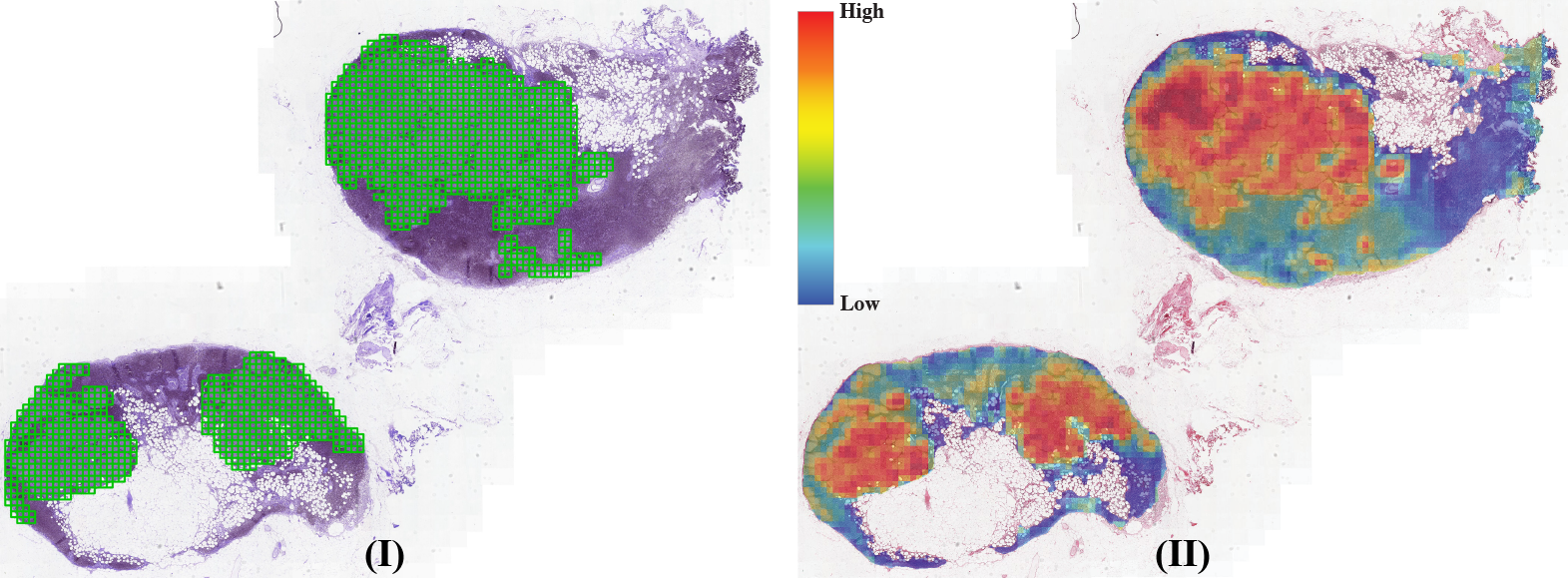}
\end{figure}

The results in Table~\ref{table2} show that  additive models do not exhibit significant improvements in classification performance, but they do show slight enhancements in reliability metrics. Notably, the simple MEAN-POOL-INS model achieves the highest reliability values, although having the lowest classification performance. \figurename~\ref{fig:cam_mean} illustrates a slide and its corresponding heatmap, highlighting the model's focus on tumor areas.

\subsubsection*{CATCH}

\begin{figure}[!ht]
	\caption{\textbf{(I) A slide from CATCH with ground truth annotations (green) overlaid on the tissue section. (II) Corresponding heatmap generated by MAX-POOL, showing predicted patch scores distribution from low (blue) to high (red).} The annotation and heatmap are spatially aligned for comparison.}
	\label{fig:cat_max}
	\includegraphics[width=\textwidth]{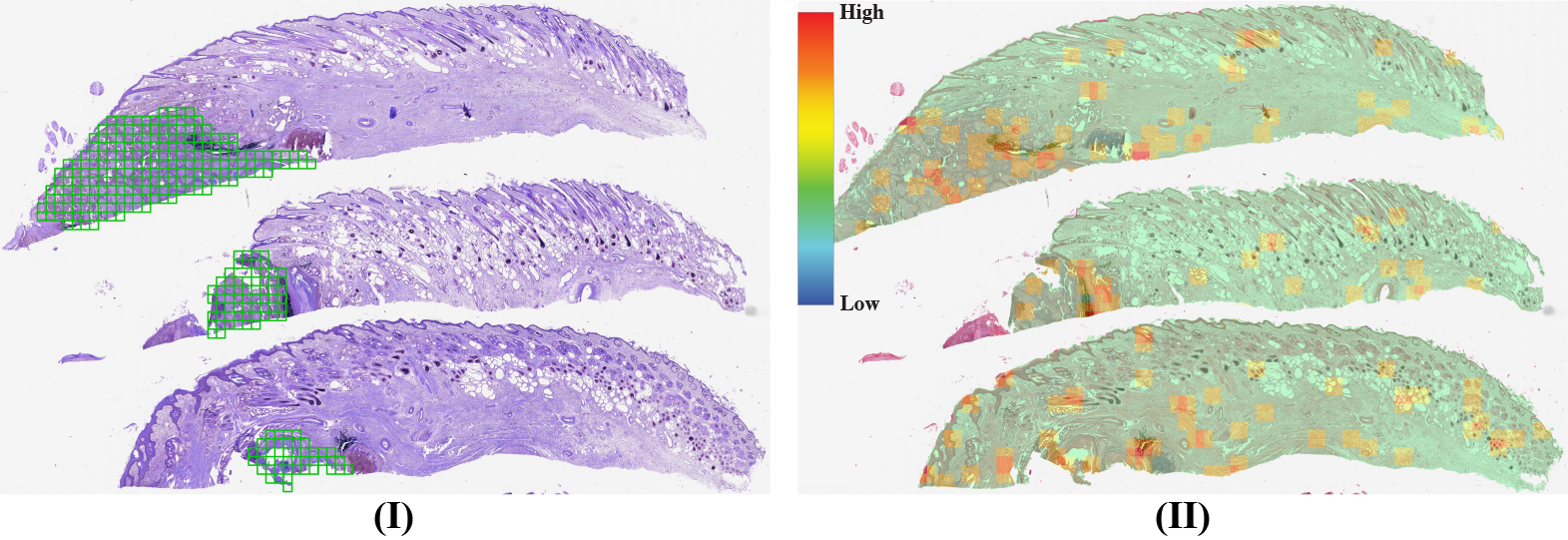}
\end{figure}

Tables~\ref{table3} and~\ref{table4} present the results for CATCH dataset. Notably, the MAX-POOL model achieves the highest classification metrics but records the lowest reliability scores. Selecting the MAX-POOL model without considering its reliability could lead to unpredictable outcomes. As shown in \figurename~\ref{fig:cat_max}, the model primarily focuses on non-tumorous regions, further highlighting its lack of reliability.

\begin{table}[!ht]
	\begin{adjustwidth}{-0.74in}{0in} 
		\centering
		\renewcommand{\arraystretch}{2} 
		
		\scriptsize
		\caption{\bf Average reliability, classification, and computational metrics (± standard deviation) over five repetitions for  CATCH. }
		\begin{tabular}{|l|c|c|c|c|c|c|c|}
			\hline
			\multirow{2}{*}{\bfseries {Model}} & \multicolumn{3}{|c|}{\bfseries Reliability} & \multicolumn{2}{|c|}{\bfseries Classification}& \multicolumn{2}{|c|}{\bfseries Computation}\\
			\cline{2-8}
			&{\bfseries MI} & \bfseries Spearman's & \bfseries AUPRC & \bfseries AUC & \bfseries F1  & \bfseries FLOPs & \bfseries Size\\
			\hline
			ABMIL & 0.30 ± 0.02	& 0.53 ± 0.01	& 0.91 ± 0.01 & 0.99 ± 0.00	 & 0.90 ± 0.02	& 94 M	& 791 K\\
			\hline
			\bfseries CLAM	&\bfseries 0.32 ± 0.01	&0.54 ± 0.00	&0.91 ± 0.00	&0.99 ± 0.00	& 0.88 ± 0.01	& 94 M	& 798 K\\
			\hline
			\bfseries DTFD	 &0.20 ± 0.06	&\bfseries 0.56 ± 0.01	&0.91 ± 0.01	&1 ± 0.00	&0.93 ± 0.01	&126 M	&1842 K\\
			\hline
			\bfseries ACMIL/4	&0.28 ± 0.01	&0.55 ± 0.01	&\bfseries 0.92 ± 0.01 	& 1 ± 0.00	&0.93 ± 0.01	&95 M	&807 K\\
			\hline
			MADMIL/2	&0.30 ± 0.01	&0.55 ± 0.00	&0.92 ± 0.00 	&0.99 ± 0.00	&0.90 ± 0.02	&79 M	&660 K\\
			\hline
			\bfseries MAX-POOL	&0.00 ± 0.00	&-0.05 ± 0.00	&0.65 ± 0.00	&\bfseries 1 ± 0.000	&\bfseries 0.94 ± 0.02	&\bfseries 63 M	&\bfseries 528 K\\
			\hline
			MEAN-POOL	&-& -&-&	0.98 ± 0.00 & 0.81 ± 0.02 &  63 M &  528 K\\
			\hline
		\end{tabular}
		\label{table3}
	\end{adjustwidth}
\end{table}

\begin{table}[!ht]
	\begin{adjustwidth}{-1.27in}{0in} 
		\centering
		\renewcommand{\arraystretch}{2} 
		
		\scriptsize
		\caption{\bf Average reliability, classification, and computational metrics (± standard deviation) over five repetitions for the CATCH using additive models.}
		\begin{tabular}{|l|c|c|c|c|c|c|c|}
			\hline
			\multirow{2}{*}{\bfseries {Model}} & \multicolumn{3}{|c|}{\bfseries Reliability} & \multicolumn{2}{|c|}{\bfseries Classification}& \multicolumn{2}{|c|}{\bfseries Computation}\\
			\cline{2-8}
			&{\bfseries MI} & \bfseries Spearman's & \bfseries AUPRC & \bfseries AUC & \bfseries F1  & \bfseries FLOPs & \bfseries Size\\
			\hline
			ABMIL-ADD-ATT& 0.29 ± 0.02	& 0.53 ± 0.00	& 0.90 ± 0.00 & \multirow{2}{*}{0.98 ± 0.00}&\multirow{2}{*}{0.90 ± 0.02}&\multirow{2}{*}{95 M} &\multirow{2}{*}{791 K} \\
			\cline{1-4}
			ABMIL-ADD-PATCH	& 0.06 ± 0.01 & 0.35 ± 0.02	& 0.83 ± 0.01 & & & &\\
			\hline
			CLAM-ADD-ATT	&0.31 ± 0.02	&0.53 ± 0.01	&0.90 ± 0.01	& \multirow{2}{*}{0.98 ± 0.00}     & \multirow{2}{*}{0.89 ± 0.02}	&	 \multirow{2}{*}{95 M }  &      \multirow{2}{*}{ 799 K}	\\
			\cline{1-4}
			CLAM-ADD-PATCH	&0.07 ± 0.00 &	0.35 ± 0.01 &	0.83 ± 0.01 & & & & \\
			\hline				
			\bfseries DTFD-PATCH	& 0.09 ± 0.00	&0.42 ± 0.01	&0.87 ± 0.00 &	\bfseries 1 ± 0.00&	\bfseries 0.93 ± 0.01&	126 M	&1842 K\\
			\hline
			\bfseries ACMIL/4-ADD-ATT	&0.23 ± 0.08	&\bfseries 0.55 ± 0.01&	 \bfseries 0.91 ± 0.01 	&\multirow{2}{*}{0.99 ± 0.00}     &\multirow{2}{*}{0.92 ± 0.01}		&\multirow{2}{*}{95 M}         &\multirow{2}{*}{806 K}	\\
			\cline{1-4}
			ACMIL/4-ADD-PATCH&	0.17 ± 0.01 &	0.44 ± 0.02	&0.88 ± 0.01&&&&\\				
			\hline
			MAX-POOL-INS&	0.00 ± 0.00 &	0.00 ± 0.01	&0.66 ± 0.00	& \multirow{2}{*} {0.99 ± 0.01} &   \multirow{2}{*} {0.89 ± 0.01}	&	\multirow{2}{*} {63 M}   &        \multirow{2}{*} {528 K}	\\
			\cline{1-4}
			MAX-POOL-INS-PATCH&	0.29 ± 0.01 &	0.47 ± 0.06	&0.88 ± 0.03&&&&\\				
			\hline
			\bfseries MEAN-POOL-INS	&\bfseries 0.32 ± 0.00&	0.45 ± 0.01	&0.85 ± 0.00 &	0.98 ± 0.00&	0.81 ± 0.02&	\bfseries 63 M&	\bfseries 528 K\\
			\hline
		\end{tabular}
		\label{table4}
	\end{adjustwidth}
\end{table}

\begin{figure}[!ht]
	\caption{\textbf{(I) A slide from CATCH with ground truth annotations (green) overlaid on the tissue section. (II) Corresponding heatmap generated by ACMIL/4, showing predicted patch scores distribution from low (blue) to high (red).} The annotation and heatmap are spatially aligned for comparison.}
	\label{fig:cat_acmil}
	\includegraphics[width=\textwidth]{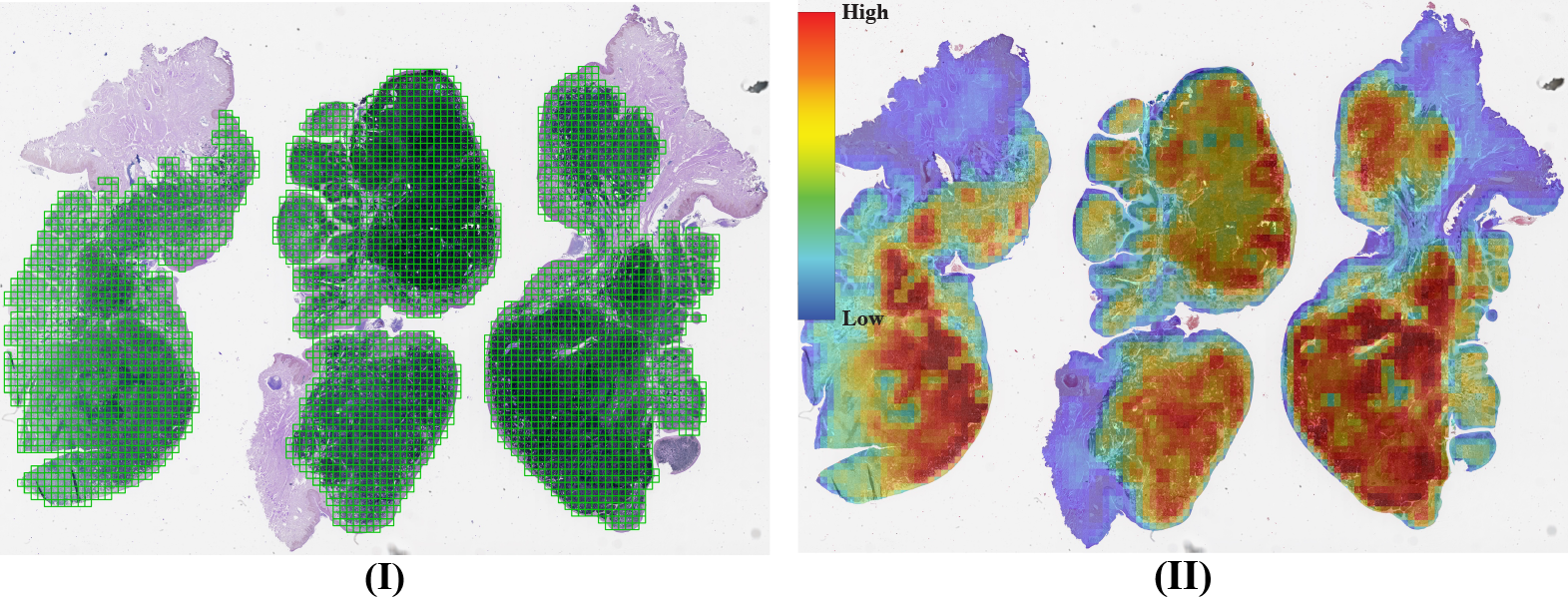}
\end{figure}

Additionally, different models excel in various reliability metrics. Among them, the ACMIL/4 models (both plain and additive) strike a good balance between classification and reliability performance, although they are computationally demanding. In \figurename~\ref{fig:cat_acmil}, the heatmap produced by the plain ACMIL/4 model demonstrates that the model accurately focuses on regions of interest as annotated by pathologists.

Finally, the MEAN-POOL-INS model, while being the most computationally efficient, exhibits the lowest F1 score but achieves moderate reliability metrics. 
				
\subsubsection*{TCGA BRCA}
The MADMIL/3 models exhibit the highest reliability performance, a high AUC classification metric, and low computational costs for this dataset, as shown in Table~\ref{table5}.

\begin{table}[!ht]
	\begin{adjustwidth}{-0.74in}{0in} 
		\centering
		\renewcommand{\arraystretch}{2} 
		
		\scriptsize
		\caption{\bf Average reliability, classification, and computational metrics (± standard deviation) over five repetitions for TCGA BRCA.}
		\begin{tabular}{|l|c|c|c|c|c|c|c|}
			\hline
			\multirow{2}{*}{\bfseries {Model}} & \multicolumn{3}{|c|}{\bfseries Reliability} & \multicolumn{2}{|c|}{\bfseries Classification}& \multicolumn{2}{|c|}{\bfseries Computation}\\
			\cline{2-8}
			&{\bfseries MI} & \bfseries Spearman's & \bfseries AUPRC & \bfseries AUC & \bfseries F1  & \bfseries FLOPs & \bfseries Size\\
			\hline
			ABMIL & 0.19 ± 0.04	& 0.44 ± 0.04	& 0.82 ± 0.02 &  	 0.95 ± 0.01& 	 0.85 ± 0.02	& 94 M	& 789 K\\
			\hline
			CLAM	& 0.18 ± 0.01 & 	0.48 ± 0.01	& 0.84 ± 0.01 & 	0.95 ± 0.01& 	0.84 ± 0.01	& 94 M& 	791 K\\
			\hline
			\bfseries DTFD	 &0.15 ± 0.03 	& 0.47 ± 0.01 	& 0.84 ± 0.01 & 	0.96 ± 0.01& 	\bfseries 0.87 ± 0.02	& 126 M	& 1053 K\\
			\hline
			ACMIL/3	& 0.22 ± 0.01 	& 0.49 ± 0.02 	& 0.85 ± 0.01	& 0.96 ± 0.01& 	 0.85 ± 0.02	& 95 M	& 792 K\\
			\hline
			\bfseries MADMIL/3	&\bfseries 0.22 ± 0.02 	& \bfseries 0.49 ± 0.03 & 	\bfseries 0.85 ± 0.02 & 	0.97 ± 0.01& 	0.85 ± 0.04	& 74 M	& 615 K\\
			\hline 
			MAX-POOL	& 0.00 ± 0.00 	& -0.06 ± 0.00 & 	0.61 ± 0.00 	& 0.93 ± 0.00	& 0.80 ± 0.02	&  63 M& 526 K	\\
			\hline
			\bfseries MEAN-POOL	&-& -&-	&  \bfseries 0.98 ± 0.01& 0.86 ± 0.02& \bfseries 63 M& \bfseries 526 K	\\
			\hline
		\end{tabular}
		\label{table5}
	\end{adjustwidth}
\end{table}

\begin{figure}[!ht]
	\caption{\textbf{(I) A slide from TCGA BRCA with ground truth annotations (green) overlaid on the tissue section. (II) Corresponding heatmap generated by MADMIL/3, showing predicted patch scores distribution from low (blue) to high (red).} The annotation and heatmap are spatially aligned for comparison.}
	\label{fig:brca_madmil}
	\includegraphics[width=\textwidth]{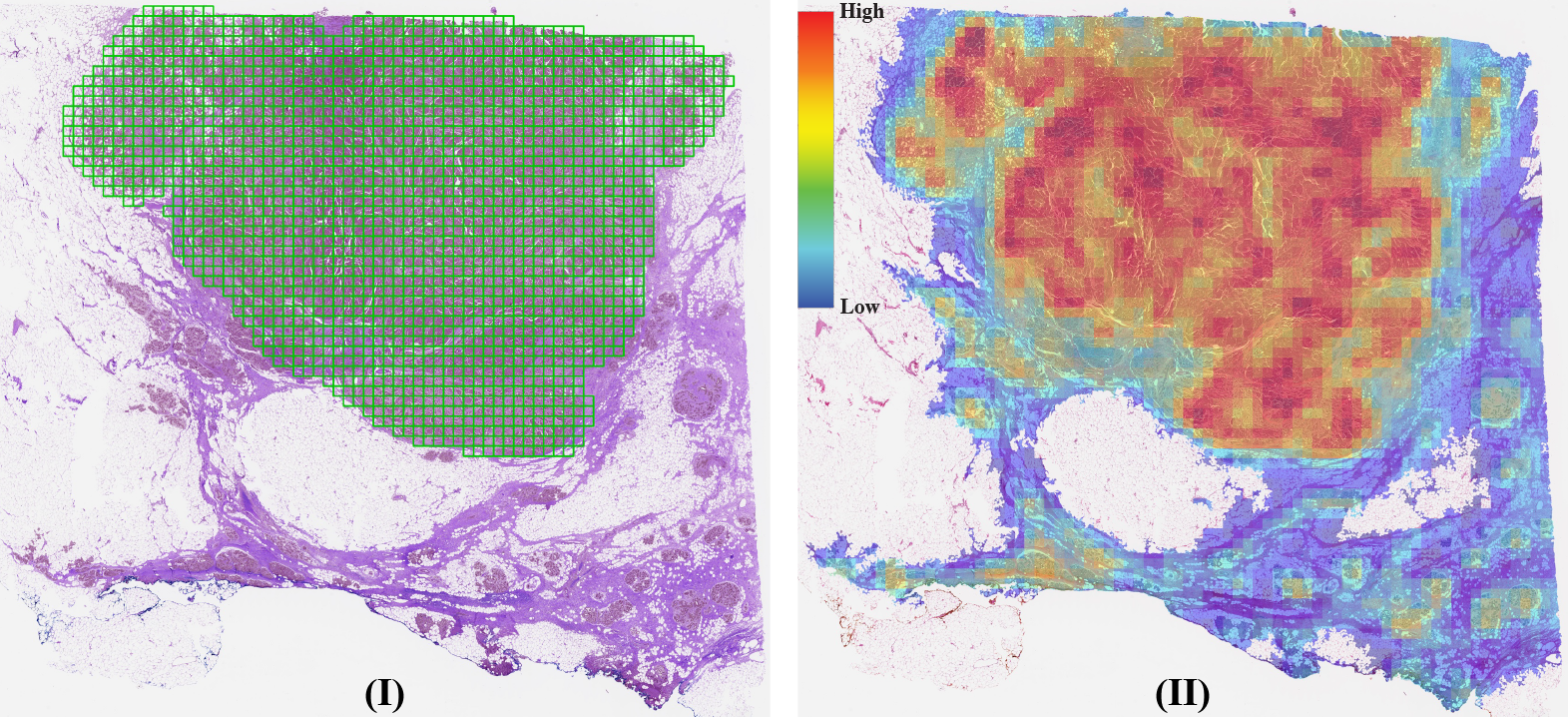}
\end{figure}

\begin{figure}[!ht]
	\caption{\textbf{(I) A slide from TCGA BRCA with ground truth annotations (green) overlaid on the tissue section. (II) Corresponding heatmap generated by MEAN-POOL-INS, showing predicted patch scores distribution from low (blue) to high (red).} The annotation and heatmap are spatially aligned for comparison.}
	\label{fig:brca_mean-pool-ins}
	\includegraphics[width=\textwidth]{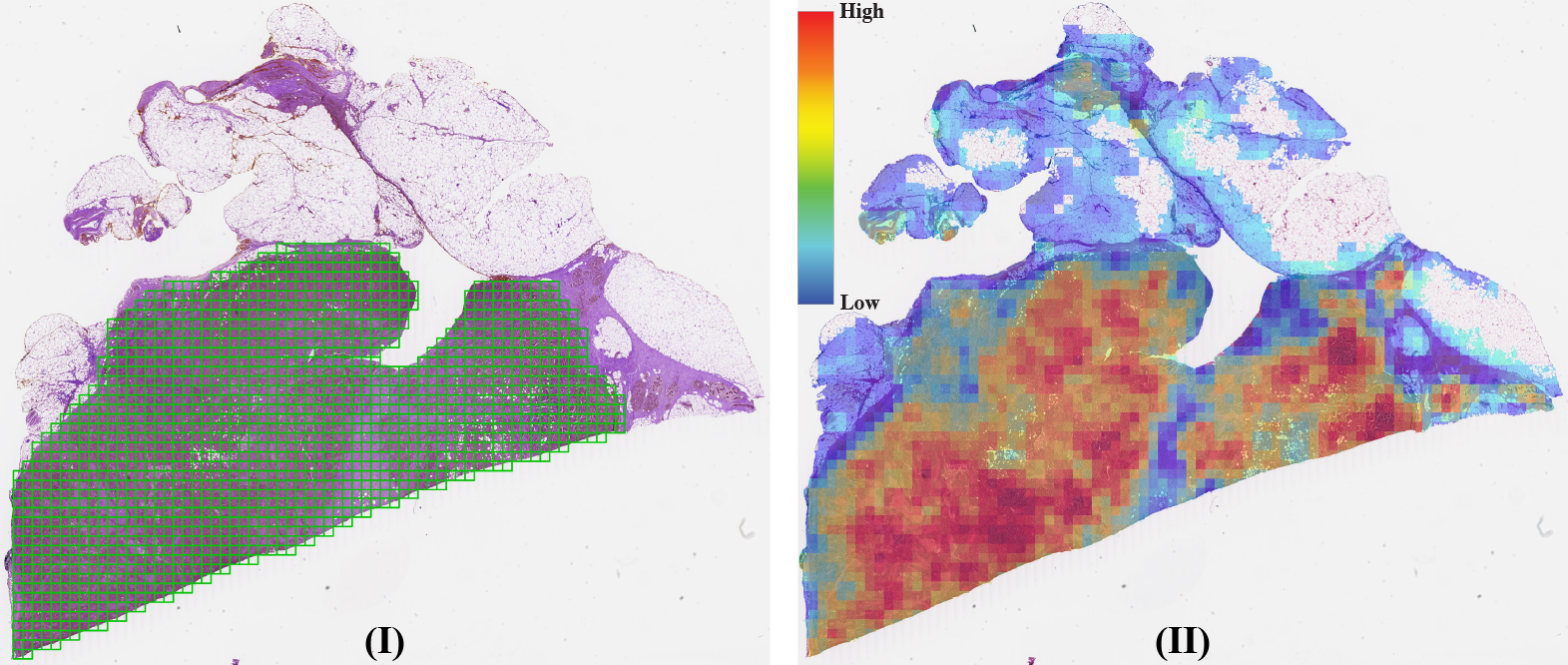}
\end{figure}

 \figurename~\ref{fig:brca_madmil} visually supports these numerical results. On the other hand, the DTFD model achieves the highest F1 score but demonstrates lower reliability, AUC, and computational efficiency. Additionally, we present a slide's heatmap generated by the MEAN-POOL-INS model in \figurename~\ref{fig:brca_mean-pool-ins}, which aligns with the model’s moderate reliability, high classification AUC, and low computational cost reported in Table~\ref{table6}.

\begin{table}[!ht]
	\begin{adjustwidth}{-1.4in}{0in} 
	\centering
	\renewcommand{\arraystretch}{2} 
	
	\scriptsize
	\caption{\bf Average reliability, classification, and computational metrics (± standard deviation) over five repetitions for TCGA BRCA using additive models.}
		\begin{tabular}{|l|c|c|c|c|c|c|c|}
				\hline
				\multirow{2}{*}{\bfseries {Model}} & \multicolumn{3}{|c|}{\bfseries Reliability} & \multicolumn{2}{|c|}{\bfseries Classification}& \multicolumn{2}{|c|}{\bfseries Computation}\\
				\cline{2-8}
				&{\bfseries MI} & \bfseries Spearman's & \bfseries AUPRC & \bfseries AUC & \bfseries F1  & \bfseries FLOPs & \bfseries Size\\
				\hline
				ABMIL-ADD-ATT& 0.19 ± 0.03	&0.41 ± 0.10&	0.81 ± 0.04&	 \multirow{2}{*} {0.91 ± 0.04} &     \multirow{2}{*} {0.80 ± 0.04}	&	\multirow{2}{*} {95 M}&	\multirow{2}{*} {789 K}	 \\
				\cline{1-4}
				ABMIL-ADD-PATCH	& 0.07 ± 0.03 &	0.33 ± 0.03 &	0.80 ± 0.01				 & & & &\\
				\hline
				\bfseries CLAM-ADD-ATT	& 0.14 ± 0.04 	 &\bfseries 0.48 ± 0.02 &	0.84 ± 0.02  &	 \multirow{2}{*} {0.92 ± 0.01} &        \multirow{2}{*} {0.79 ± 0.06}	 &	\multirow{2}{*} {95 M} &	\multirow{2}{*} {791 K}	\\
				\cline{1-4}
				CLAM-ADD-PATCH	& 0.06 ± 0.02	 &0.29 ± 0.04 &	0.79 ± 0.00 				 & & & & \\
				\hline				
				\bfseries DTFD-PATCH	& 0.07 ± 0.01	&0.29 ± 0.02&	0.79 ± 0.01 	&0.96 ± 0.01	&\bfseries 0.87 ± 0.02	&126 M	&1053 K\\
				\hline
				ACMIL/3-ADD-ATT	&  0.12 ± 0.01	&0.23 ± 0.30&	0.73 ± 0.12	&\multirow{2}{*} {0.94 ± 0.01}  &   \multirow{2}{*} {0.84 ± 0.02}	&	\multirow{2}{*} {95 M}	&\multirow{2}{*} {792 K}	\\
				\cline{1-4}
				ACMIL/3-ADD-PATCH&	0.06 ± 0.01	&0.11 ± 0.08 &	0.70 ± 0.04 					&&&&\\				
				\hline
				MAX-POOL-INS&	0.00 ± 0.00	&0.01 ± 0.01	&0.61 ± 0.00	&\multirow{2}{*} {0.89 ± 0.04}  &  \multirow{2}{*} {0.73 ± 0.06}		&\multirow{2}{*} {63 M}	&\multirow{2}{*} {526 K}	\\
				\cline{1-4}
				\bfseries MAX-POOL-INS-PATCH&	0.24 ± 0.04 &	0.45 ± 0.04 	& \bfseries 0.84 ± 0.02 				&&&&\\				
				\hline
				\bfseries MEAN-POOL-INS	& \bfseries 0.27 ± 0.00&	 0.30 ± 0.01 &	0.79 ± 0.01 &	\bfseries 0.98 ± 0.01&	0.85 ± 0.02	&\bfseries 63 M	& \bfseries 526 K\\
				\hline
		\end{tabular}
		\label{table6}
\end{adjustwidth}
\end{table}

\section*{Discussion}
\begin{figure}[!ht]
	\centering
	\caption{\textbf{Bar plots comparing the models with different metrics.}}
	\label{fig:bar_plot}
	\includegraphics[width=\linewidth]{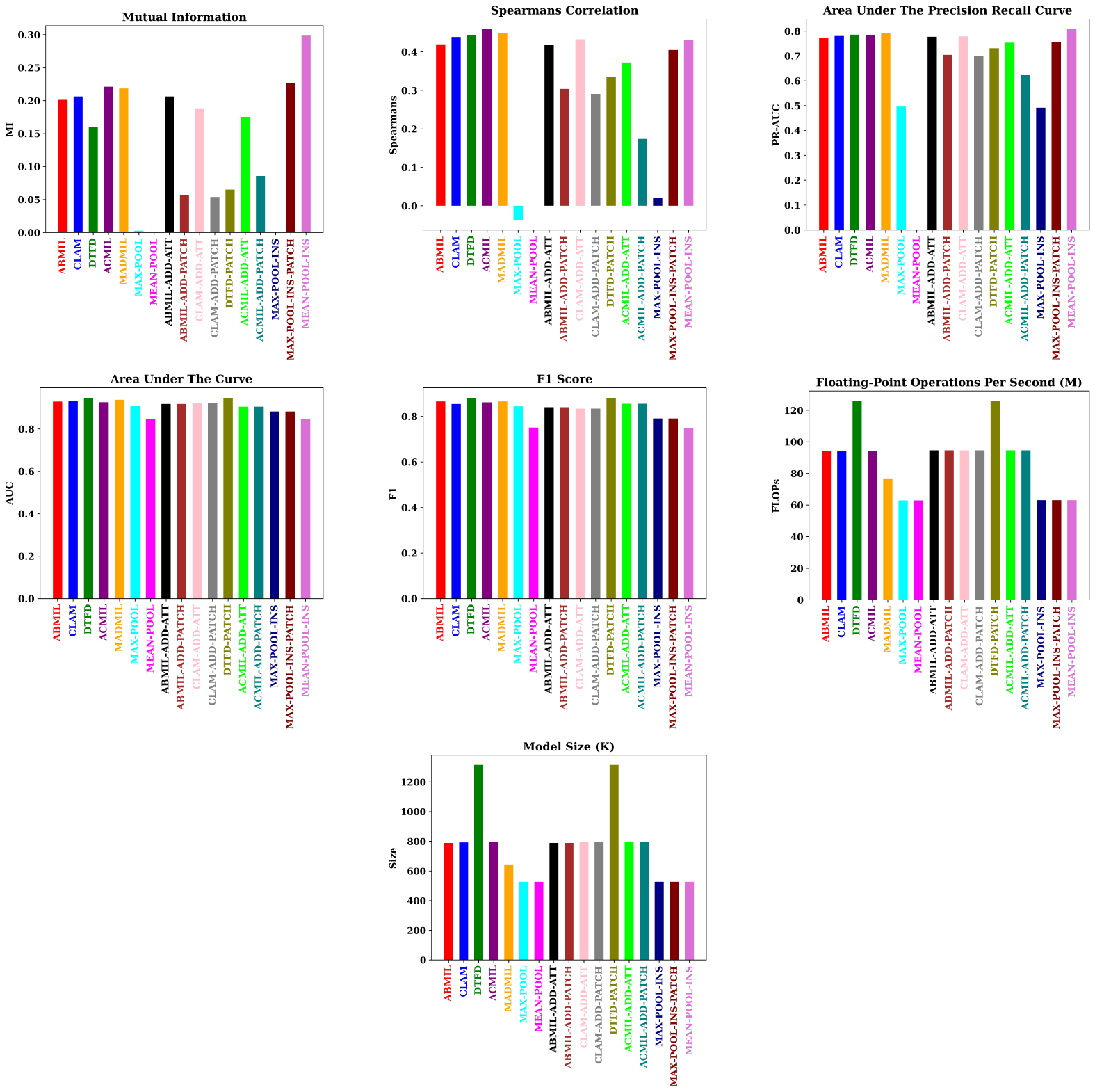}
\end{figure}

Based on the overall mean of the average results across the three datasets, as presented in Tables~\ref{table7},~\ref{table8}, and \figurename~\ref{fig:bar_plot}, several key conclusions can be drawn. Repeated-measures ANOVA revealed significant differences across models for all evaluated metrics. For each model the three datasets were treated as repeated measures.  Specifically, significant effects of model were observed on MI ($F(5,20) = 178.47$, $p < 0.001$), Spearman correlation ($F(5,20) = 12.34$, $p = 0.0001$), AUPRC ($F(5,20) = 9.12$, $p = 0.0005$), AUC ($F(6,24) = 15.21$, $p < 0.001$), and F1 score ($F(6,24) = 8.45$, $p = 0.0012$), indicating that model performances differed significantly across datasets. The learning curves for all methods on the TCGA BRCA dataset are provided in the Supporting information. These curves illustrate the convergence dynamics and stability across folds, supporting the reported performance metrics.

The model comparisons highlight important trade-offs. First, while the simple MAX-POOL network achieves high classification metrics and low computational cost, it demonstrates poor reliability values. This inconsistency may compromise clinical trust, suggesting that despite its efficiency, MAX-POOL is unsuitable for deployment in real-world applications where reliability is critical. In contrast, the MEAN-POOL-INS model shows high reliability, comparatively good classification performance, and low computational cost, making it a more favorable choice for scenarios where consistent predictions and resource constraints are both important.

\begin{table}[!ht]
	\centering
	\renewcommand{\arraystretch}{2} 
	
	\scriptsize
	\caption{\bf The overall mean of the average reliability, classification, and computation metrics across CAMELYON16, CATCH, and TCGA BRCA.}
		\begin{tabular}{|l|c|c|c|c|c|c|c|}
			\hline
			\multirow{2}{*}{\bfseries {Model}} & \multicolumn{3}{|c|}{\bfseries Reliability} & \multicolumn{2}{|c|}{\bfseries Classification} & \multicolumn{2}{|c|}{\bfseries Computation} \\
			\cline{2-8}
			& {\bfseries MI} & \bfseries Spearman's & \bfseries AUPRC & \bfseries AUC & \bfseries F1  & \bfseries FLOPs & \bfseries Size\\
			\hline
			ABMIL & 0.20& 0.42&0.77& 0.93 &0.87 & 94 M & 790 K \\
			\hline
			CLAM & 0.21 &0.44 & 0.78& 0.93& 0.86& 94 M & 793 K \\
			\hline
			\bfseries  DTFD & 0.16& 0.44& 0.79& \bfseries  0.95&  \bfseries 0.88& 126 M & 1316 K \\
			\hline
			\bfseries  ACMIL & \bfseries 0.22 & \bfseries  0.46& 0.78&0.93& 0.86& 94 M & 797 K \\
			\hline
			\bfseries  MADMIL & 0.22& 0.45  &  \bfseries  0.79 & 0.94& 0.87 & 77 M & 644 K \\
			\hline
			\bfseries MAX-POOL & 0.00 & -0.04& 0.50 & 0.91& 0.84 & \bfseries 63 M & \bfseries  527 K \\
			\hline
			MEAN-POOL & - & - & - & 0.85  & 0.75  & 63 M & 527 K \\
			\hline
		\end{tabular}
\label{table7}
\end{table}

\begin{table}[!ht]
		\begin{adjustwidth}{-0in}{0in} 
		\centering
		\renewcommand{\arraystretch}{2} 
		
		\scriptsize
		\caption{\bf The overall mean of the average reliability, classification, and computation metrics  for additive models across CAMELYON16, CATCH, and TCGA BRCA datasets.}
		\begin{tabular}{|l|c|c|c|c|c|c|c|}
			\hline
			\multirow{2}{*}{\bfseries Model} & \multicolumn{3}{|c|}{\bfseries Reliability} & \multicolumn{2}{|c|}{\bfseries Classification} & \multicolumn{2}{|c|}{\bfseries Computation} \\
			\cline{2-8}
			& \bfseries MI & \bfseries Spearman's & \bfseries AUPRC & \bfseries AUC & \bfseries F1 & \bfseries FLOPs & \bfseries Size \\
			\hline
			ABMIL-ADD-ATT & 0.21 & 0.42& 0.78 &  \multirow{2}{*} {0.92} &  \multirow{2}{*} {0.84} &  \multirow{2}{*} {95 M} &  \multirow{2}{*} {790 K} \\
			\cline{1-4}
			ABMIL-ADD-PATCH & 0.06& 0.30& 0.70 & & & &\\
			\hline
			\bfseries CLAM-ADD-ATT & 0.19 & \bfseries 0.43& 0.78 & \multirow{2}{*} {0.92} & \multirow{2}{*} {0.83} & \multirow{2}{*} {95 M} & \multirow{2}{*} {793 K} \\
			\cline{1-4}
			CLAM-ADD-PATCH & 0.05& 0.29& 0.70& & & &\\
			\hline
			\bfseries DTFD-PATCH & 0.07 & 0.33& 0.73& \bfseries 0.95 & \bfseries 0.88& 126 M & 1316 K \\
			\hline
			ACMIL-ADD-ATT & 0.18 & 0.37& 0.75& \multirow{2}{*} {0.91} & \multirow{2}{*} {0.86} & \multirow{2}{*} {95 M} & \multirow{2}{*} {797 K} \\
			\cline{1-4}
			ACMIL-ADD-PATCH & 0.09 & 0.17& 0.62& & & &\\
			\hline
			MAX-POOL-INS & 0.00& 0.02& 0.49& \multirow{2}{*} {0.88} & \multirow{2}{*} {0.79} & \multirow{2}{*} {63 M} & \multirow{2}{*} {527 K} \\
			\cline{1-4}
			MAX-POOL-INS-PATCH & 0.23 & 0.40& 0.76& & & &\\
			\hline		
			\bfseries MEAN-POOL-INS & \bfseries 0.30&  0.43& \bfseries 0.81& 0.85& 0.75& \bfseries 63 M & \bfseries 527 K \\
			\hline
		\end{tabular}
		\label{table8}
	\end{adjustwidth}
\end{table}

The multihead networks, ACMIL and MADMIL, offer both high reliability and classification performance, but at a higher computational expense, highlighting their potential for research or clinical applications where both accuracy and consistency are essential. Their higher computational cost, particularly for ACMIL, indicates a trade-off that must be considered in resource-limited settings. Meanwhile, the DTFD model delivers the highest classification performance, though this comes at the cost of increased computational demands and reduced reliability. Therefore, when classification performance is a priority, and computational cost and reliability are secondary concerns, DTFD may be a suitable choice. 

Regarding the additive models, while they do not exhibit a significant improvement in classification metrics, they do show a subtle increase in reliability scores, albeit with slightly higher computational cost. While they offer enhanced interpretability, their limited benefits in reliability and accuracy indicate that they may be most valuable in research contexts emphasizing model explainability rather than operational performance.

To better understand the contribution of each reliability metric, we examined the model performance by selectively excluding MI, Spearman correlation, or AUPRC from the overall model ranking (Table~\ref{table9}). The results show that excluding MI substantially reduces the ability to differentiate models with consistent predictions, as several top models become tied, highlighting MI’s critical role in capturing prediction consistency. Excluding AUPRC also affects the rankings, emphasizing its importance in assessing realiability. Similarly, removing Spearman correlation impacts the overall ranking, causing several models to tie at lower positions. These findings indicate that all three metrics contribute meaningfully to assessing model reliability, and a multi-metric framework is necessary to capture different aspects of predictive behavior effectively.

\begin{table}[!ht]
	\begin{adjustwidth}{-1.4in}{0in} 
	\centering
	\renewcommand{\arraystretch}{2}
	\scriptsize
	\caption{\textbf{Analysis of reliability metrics showing the effect of excluding each metric on model ranking.} Rankings are obtained by summing the scores of the selected reliability metrics (with equal weight) across CAMELYON16, CATCH, and TCGA BRCA datasets, and ranking models based on the aggregated score.}
	
	\begin{tabular}{|l|c|}
		\hline
		\textbf{Metric Excluded} & \textbf{Model Ranking (Reliability)}\\
		\hline
		None (All metrics) & MEAN-POOL-INS $>$ ACMIL $>$ MADMIL $>$ CLAM $>$ ABMIL-ADD-ATT $>$ CLAM-ADD-ATT\\
		\hline
		MI excluded & MEAN-POOL-INS $=$ ACMIL $=$ MADMIL $>$ DTFD $>$ CLAM $>$ CLAM-ADD-ATT \\
		\hline
		Spearman's excluded & MEAN-POOL-INS $>$ MADMIL $>$ ACMIL $>$ CLAM $=$ ABMIL-ADD-ATT $=$ MAX-POOL-INS-PATCH \\
		\hline
		AUPRC excluded & MEAN-POOL-INS $>$ ACMIL $>$ MADMIL $>$ CLAM $>$ ABMIL-ADD-ATT $=$ MAX-POOL-INS-PATCH\\
		\hline
	\end{tabular}
		\label{table9}
	\end{adjustwidth}
\end{table}

Overall, these findings underscore that classification metrics alone do not capture the full utility of a model. Incorporating reliability and computational cost alongside accuracy provides a more comprehensive assessment, guiding model selection. While standard metrics such as AUC, accuracy, and F1 capture overall classification performance, they can fail in certain scenarios, such as rare positive instances, highly imbalanced classes, or heterogeneous MIL bags. In our experiments, the MAX-POOL network achieves high AUC and F1 scores but demonstrates low reliability, highlighting that strong classification metrics alone do not guarantee trustworthy predictions. Incorporating reliability metrics alongside standard metrics allows identification of models that not only perform well on average but also make consistent and robust predictions, providing a more comprehensive assessment of model behavior and guiding informed model selection. By explicitly considering these factors, researchers can design MIL architectures that balance predictive performance, computational efficiency, and deployability, ultimately advancing the development of models suited for real-world WSI analysis.

\section*{Conclusion}

In this paper, we proposed a novel approach to comparing MIL models in terms of reliability using three distinct metrics. Notably, we found that the MEAN-POOL-INS model achieves high reliability despite its simple architecture and low computational cost. Additionally, the multihead models, while slightly more computationally expensive, demonstrate higher classification performance and maintain fair reliability.

We hope that future research will incorporate reliability and computational metrics, leading to the development of more dependable and efficient models for WSI classification.

\section*{Supporting information}

	\begin{figure}[!ht]
	\centering
	\caption*{\textbf{S1 Fig. Learning curve of the models (set 1).}  Each panel shows the training curve for one model on TCGA BRCA.}
	
	\label{fig:s1}
	\includegraphics[width=\textwidth]{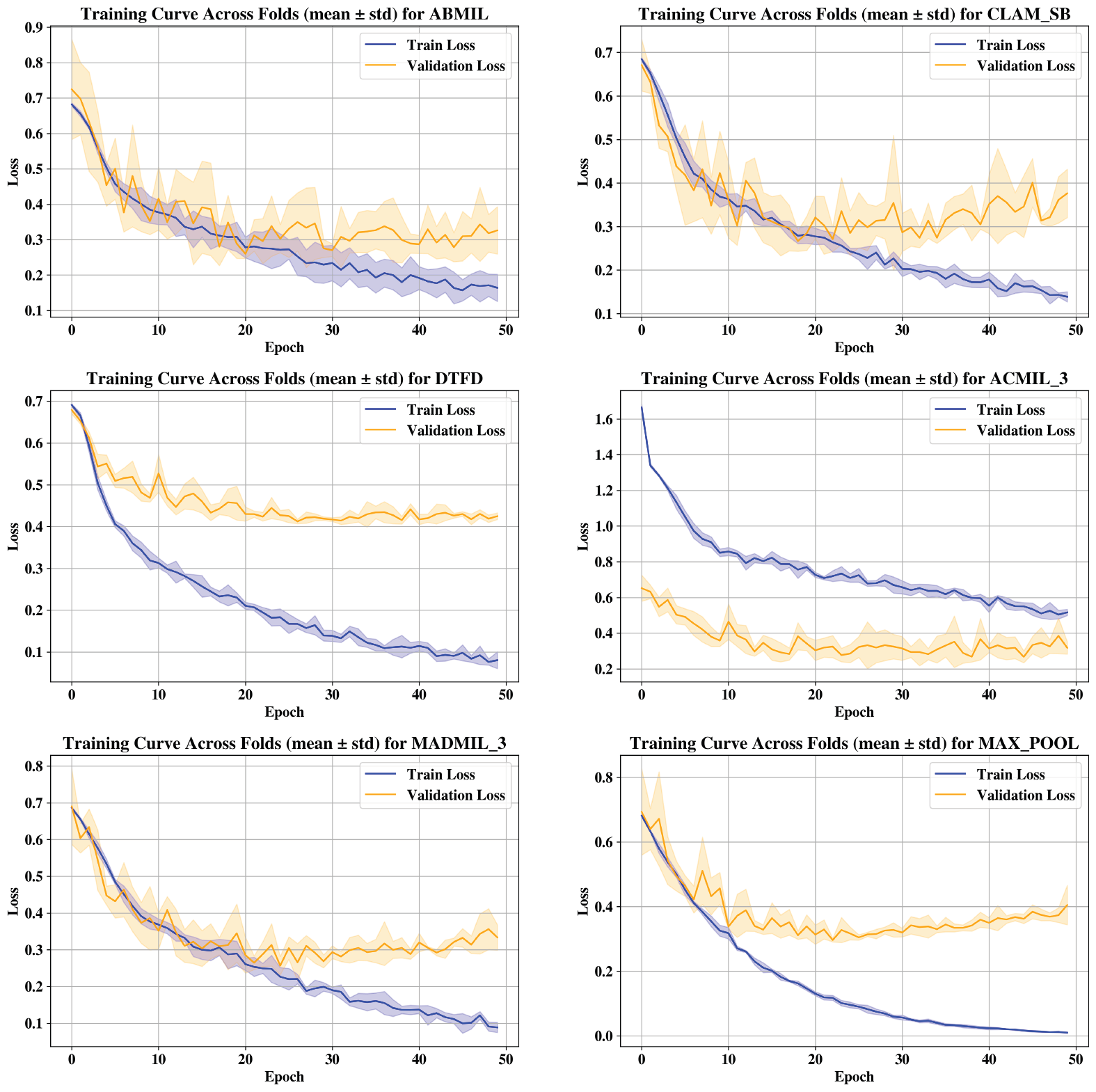}
\end{figure}

\begin{figure}[!ht]
	\centering
	\caption*{\textbf{S2 Fig. Learning curve of the models (set 2).}  Each panel shows the training curve for one model on TCGA BRCA.}
	\label{fig:s2}
	\includegraphics[width=\textwidth]{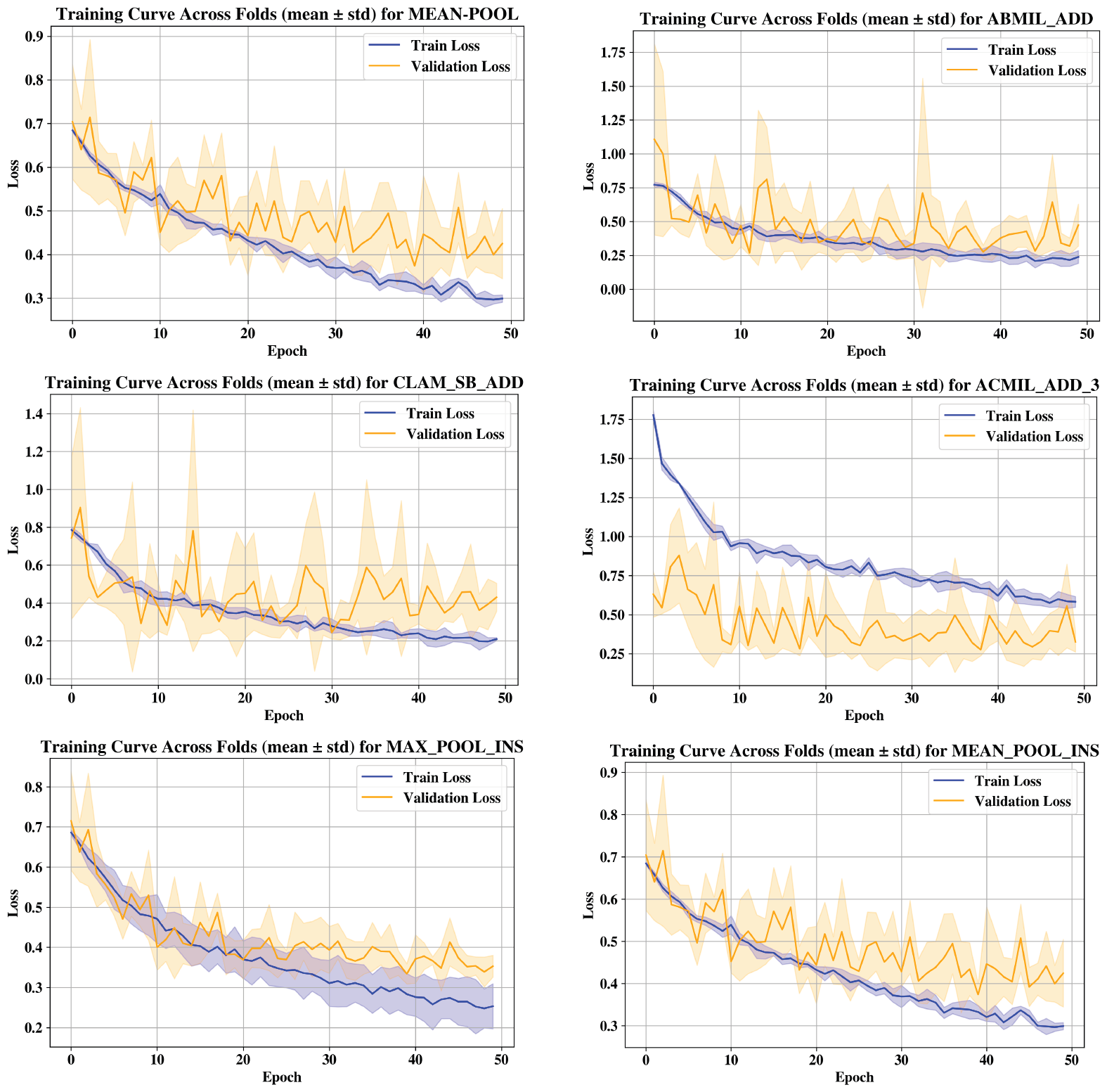}
		
\end{figure}

\section*{Data Availability Statement}
The code for reproducing our results is available at  \hyperlink{https://github.com/tueimage/MIL-Reliability}{https://github.com/tueimage/MIL-Reliability}.

\section*{Competing interests}
The authors have declared that no competing interests exist. This does not alter our adherence to PLOS ONE policies on sharing data and materials.

\section*{Author Contributions}
\renewcommand{\arraystretch}{1.2}
\begin{tabularx}{\textwidth}{@{}l X@{}}
	Conceptualization: & Hassan Keshvarikhojasteh, Mitko Veta \\
	Data curation: & Marc Aubreville, Christof A. Bertram \\
	Formal analysis: & Hassan Keshvarikhojasteh \\
	Methodology: & Hassan Keshvarikhojasteh, Mitko Veta \\
	Supervision: & Mitko Veta, Josien P.W. Pluim \\
	Writing – original draft: & Hassan Keshvarikhojasteh \\
	Writing – review \& editing: & Hassan Keshvarikhojaste, Marc Aubreville, Christof A. Bertram, Josien P.W. Pluim, Mitko Veta
\end{tabularx}

\section*{Funding}

This research was supported by the IMI BigPicture project (IMI945358), part of the Innovative Medicines Initiative 2 Joint Undertaking under grant agreement No 945358. The funders had no role in study design, data collection and analysis, decision to publish, or preparation of the manuscript.

\nolinenumbers

\end{document}